%% file: main.tex
\documentclass[letterpaper]{article} 
\usepackage[preprint]{aaai2027}      
\usepackage[hyphens]{url}  
\usepackage{graphicx} 
\usepackage{natbib}  
\usepackage{caption} 
\usepackage{booktabs}
\usepackage{multirow}
\usepackage{amsmath}
\usepackage{amssymb}

\title{TangPoetryBench: A Multi-Dimensional Benchmark and Rubric-Conditioned Evaluator for Poetry-to-Image Generation}

\author{
    Haoqi Hu\corresponding,
    Tongji Luo,
    Li Zhang,
    Boning Zhou
}
\affiliations{
    Independent Researcher\\
    williamhu919@gmail.com, luotongji@gmail.com, zhangli.stark@gmail.com, boningzhou0731@gmail.com
}

\begin{document}
\maketitle

\begin{abstract}
Text-to-image (T2I) models are increasingly asked to illustrate literary and cultural
content, yet we cannot measure how well an image renders the \emph{meaning} of a poem. The
task is many-sided: a good illustration must be visually sound, faithful to the poem's
imagery and scene, culturally and stylistically appropriate, free of spurious text, and true to its
emotion, and its deepest requirements, imagery and especially implicit emotion, are never
stated in the words. Existing metrics (CLIPScore, BLIPScore, VQAScore) reward literal
text-image correspondence and so cannot tell whether an illustration succeeds, let alone
why, or even separate the best model from the worst. We introduce \textbf{TangPoetryBench},
a multi-dimensional benchmark of 1{,}280 images (320 classical Chinese Tang poems $\times$ 4
state-of-the-art T2I models) with quality-controlled human annotations across ten
dimensions. Analyzing this data, we reveal the shared and model-specific strengths and
weaknesses of current T2I models, including their ability to evoke a poem's implicit
emotion. We further introduce
\textbf{PoemAutoEvaluator (PAE)}, an open, rubric-conditioned evaluator that reaches parity
with a strong proprietary judge (Claude), generalizes to an unseen generator and a second
poetic tradition (Song Ci), and lets the benchmark scale to new images without fresh human
annotation. We release the benchmark, annotations, and evaluator.
\end{abstract}


\input{sections/introduction}
\input{sections/related_work}
\input{sections/benchmark}
\input{sections/analysis}
\input{sections/pae}
\input{sections/conclusion}

\typeout{REFPAGE=\thepage}
\bibliography{main}

\clearpage
\appendix
\section*{Appendix}
\suppressfloats[t] 
\input{sections/appendix}

\end{document}

%% file: sections/introduction.tex
\section{Introduction}
\label{sec:intro}

Classical Chinese Tang poetry is among the most widely read literary traditions in the
world: its canon is memorized by schoolchildren and has inspired painting for over a
millennium. A single quatrain compresses concrete imagery, seasonal cues, historical
allusion, and emotional undertone into a few lines. Illustrating such a poem is a
demanding test of whether a generative model can move past literal prompt-following to
render content that is implicit, metaphorical, and affective. Modern text-to-image (T2I)
models \citep{betker2023improving,rombach2022high} produce strikingly detailed images
from a prompt and are increasingly applied to literary and cultural illustration, yet our
ability to \emph{evaluate} whether such an illustration captures a poem has not kept pace:
poetry illustration is increasingly produced, but its evaluation is still left to generic
image-text metrics never designed for it, with no benchmark built for the task.

\begin{figure}[t]
\centering
\begin{minipage}[t]{0.40\columnwidth}
\vspace{0pt}
\includegraphics[width=\linewidth]{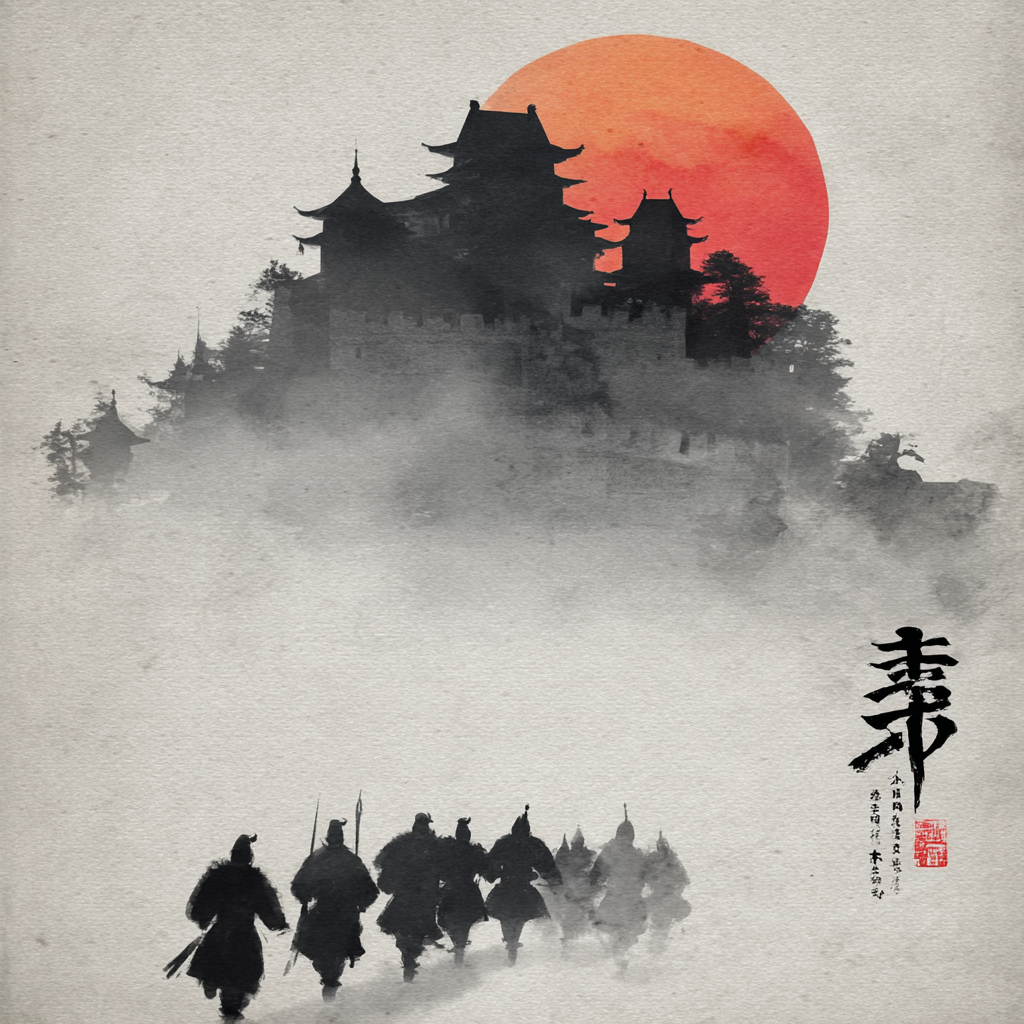}
\end{minipage}\hfill
\begin{minipage}[t]{0.57\columnwidth}
\vspace{0pt}
\includegraphics[width=\linewidth]{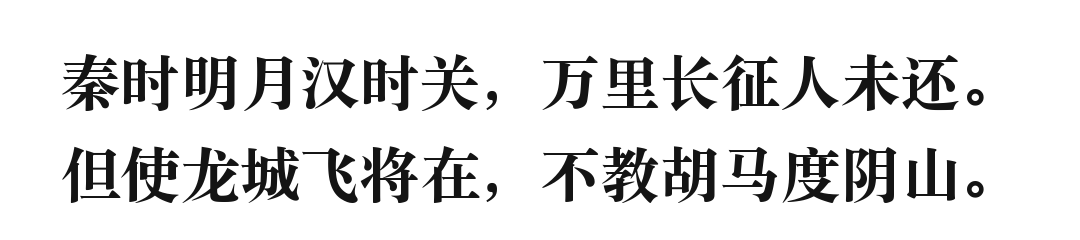}\\[4pt]
{\footnotesize\itshape The moon of Qin, the passes of Han; ten thousand \emph{li}, the
campaigners not yet returned. Were the Flying General of Dragon City still here, no Hu
horsemen would cross the Yin Mountains.}
\end{minipage}\\[6pt]
\begin{minipage}[t]{0.40\columnwidth}
\vspace{0pt}
\includegraphics[width=\linewidth]{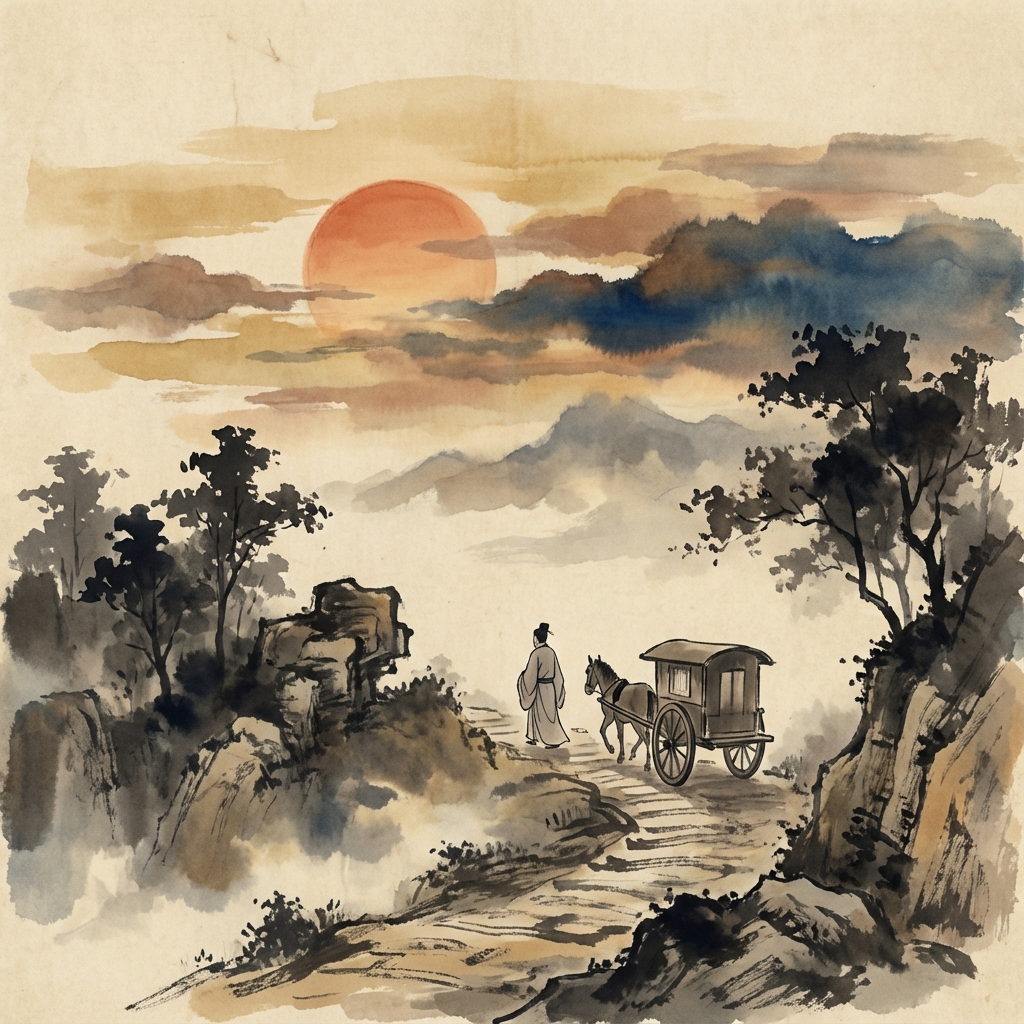}
\end{minipage}\hfill
\begin{minipage}[t]{0.57\columnwidth}
\vspace{0pt}
\includegraphics[width=\linewidth]{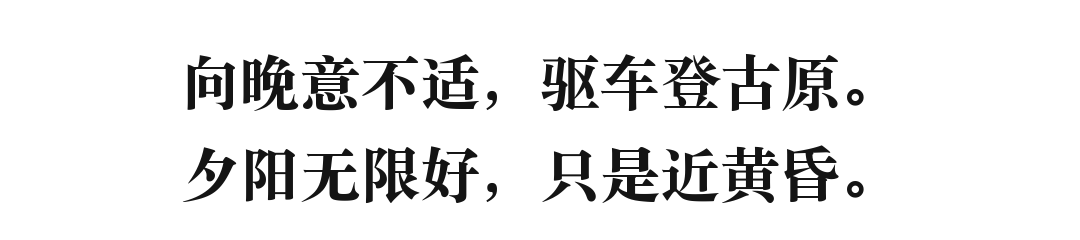}\\[4pt]
{\footnotesize\itshape At dusk, my heart uneasy, I drive up to the ancient plain. The
setting sun is boundlessly lovely, only it is near nightfall.}
\end{minipage}
\caption{Illustrating a poem demands many things at once. \textbf{Top} (Wang Changling,
``Over the Border''): a polished, beautiful image that still fails, it renders a generic
landscape and misses the frontier poem's martial desolation, so craft alone is not a good
match. \textbf{Bottom} (Li Shangyin, ``Climbing the Leyou Plateau''): a successful
illustration that captures the poem's imagery, dusk setting, and melancholy together. A
good illustration must satisfy faithful depiction, cultural and aesthetic fit, and
emotional resonance at once; being beautiful is not enough.}
\label{fig:complexity}
\end{figure}

The core obstacle is that a poem's meaning lies far beyond its literal words. Take a
well-known frontier poem (Figure~\ref{fig:complexity}, top): its lines name only the moon
over the passes and soldiers marching ten thousand \emph{li}, yet the poem is really about
the desolation of endless war and the longing for home. A poem thus has two layers: the
objects it names (the moon, the frontier passes) and the emotion it carries (the weariness
of unending campaigns). A good illustration must convey the emotion, not just draw the
objects.

These two layers can come apart, and that is what defeats standard metrics. An image can
render the frontier landscape yet convey none of its desolation, or evoke the feeling with
few of the named objects. CLIPScore \citep{hessel2021clipscore}, a BLIP matching score
\citep{li2022blip}, and even the stronger VQAScore \citep{lin2024vqascore} score only the
first layer, reducing an image to a single number for how well its objects match the words.
Such a metric rewards the literal but lifeless scene and penalizes the image that truly
captures the poem, the opposite of human judgment. It cannot even tell \emph{whether} an
illustration succeeds, let alone \emph{why} it fails.

Objects and emotion are the sharpest divide, but they are not the whole of it. A good
illustration must be visually sound, faithfully depict the poem's imagery and scene,
respect its cultural and historical setting, adopt a fitting style, avoid spurious text,
and convey the right emotion (Figure~\ref{fig:complexity}). Judging all of these requires
people, not a single alignment score. We build
\textbf{TangPoetryBench}, a benchmark of 1{,}280 images with quality-controlled human
ratings across these dimensions, and use it two ways. First, the aggregated human data
reveals what current T2I models can and cannot do with poetry, a finding about the world
independent of any model we train: they handle a poem's visual surface but fall short of
its deeper meaning, with failures that differ sharply across models.
Second, we train an automatic evaluator that replicates these human judgments so they
scale to new images and generators without fresh annotation.

Our contributions are:
\begin{itemize}
\item \textbf{TangPoetryBench}, to our knowledge the first multi-dimensional,
human-grounded benchmark built specifically for poetry-to-image generation: 1{,}280 images
with quality-controlled human annotations spanning visual quality, faithful depiction of
imagery and scene, cultural and historical correctness, artistic style, text integrity,
and emotional resonance, plus a poem-recognizability track and a hand-verified
text-integrity label, all with a transparent, reproducible adjudication procedure.
\item \textbf{A diagnostic analysis} of poetry illustration, organized around the task
rather than a leaderboard: the depict-to-evoke gap, the finding that
quality is driven by imagery and emotion rather than polish, the finding that abstraction
(not length) determines difficulty, and distinct per-model failure signatures.
\item \textbf{PoemAutoEvaluator (PAE)}, a rubric-conditioned evaluator that scores images
against a written rubric, replicates human per-dimension judgment, and generalizes to an
unseen generator; its rubric conditioning makes extension to new dimensions and
traditions a matter of supplying a rubric, not retraining an architecture.
\end{itemize}

%% file: sections/related_work.tex
\section{Related Work}
\label{sec:related}

Evaluation of text-to-image (T2I) generation has moved from single distributional scores
such as FID \citep{heusel2017fid} toward structured, multi-aspect benchmarks grounded in
human ratings. Compositional suites (DrawBench \citep{saharia2022photorealistic},
PartiPrompts \citep{yu2022scaling}, T2I-CompBench \citep{huang2023t2icompbench},
GenAI-Bench \citep{li2024genai}) probe attribute binding, spatial relations, and counting,
while HEIM \citep{lee2024heim} and EvalMuse-40k \citep{han2024evalmuse} broaden coverage to
many aspects at scale. This line establishes the value of fine-grained, multi-dimensional
evaluation, the stance we adopt. Their prompts, however, are largely explicit and compositional:
they mainly test whether named objects and relations appear, not whether an image conveys meaning
that the text only implies.

The metrics underlying these benchmarks share that assumption. Similarity- and
matching-based scores (CLIPScore \citep{hessel2021clipscore}, a BLIP matching score
\citep{li2022blip}), question-answering
scores (VQAScore \citep{lin2024vqascore}, DSG \citep{cho2023winogroundt2i}), and image-quality
predictors \citep{ke2021musiq,wang2023clipiqa} each reduce an image to a
scalar of literal correspondence or surface appeal. As we show
(Section~\ref{sec:beyond}), such a score cannot tell whether an illustration captures a
poem, and being one-dimensional it cannot say why one fails or succeeds. This motivates both human
annotation and a multi-dimensional learned judge.

A parallel line targets cultural competence. CUBE \citep{kannen2024cube}, CULTIVate
\citep{nag2025cultivate}, and analyses of the cultural gap in T2I \citep{liu2023c3} test
whether models render culture-specific artifacts, and for Chinese content CII-Bench
\citep{zhao2024ciibench} and TCC-Bench \citep{wang2025tccbench} probe multimodal cultural
understanding. These are close in spirit to our focus on a single tradition, but they
check the presence of literal cultural descriptors or test comprehension of existing
images; none asks whether a generated image evokes the implicit, affective meaning of a
poem.

Learned reward models fit a scalar to human preference (ImageReward
\citep{xu2023imagereward}, PickScore \citep{kirstain2023pickapic}, HPSv2
\citep{wu2023hpsv2}); they improve on embedding similarity but still output a single
preference number, not the per-dimension, rubric-grounded judgment we need. Closest to our
evaluator, multimodal models are increasingly used as open-ended image judges
\citep{ku2024viescore}. In parallel, poetry-to-image research
\citep{wang2025poeticvisions,yadav2025poetrytopixel,yadav2025p4i} produces illustrations
but scores them largely with generic alignment metrics that, as we demonstrate, fail for poetry
illustration. No
prior work brings these threads together: a multi-dimensional, human-grounded benchmark for
poetry illustration \emph{and} a trained, rubric-conditioned evaluator that replicates
human judgment. That is the gap we fill.


%% file: sections/benchmark.tex
\section{The TangPoetryBench Benchmark}
\label{sec:benchmark}

\paragraph{Poems and images.}
We curate 320 poems from the canonical anthology \emph{Three Hundred Tang Poems},
spanning themes (landscape, farewell, war, reflection), major poets, and forms (quatrains,
regulated verse, ancient-style verse). Each poem is illustrated by four state-of-the-art
T2I models, Midjourney V7 (MJ), Google Nano Banana Pro (Nano), OpenAI \texttt{gpt-image-1}
(GPT), and ByteDance Seedream 4.5 (Seedream), giving 1{,}280 images. All models receive an
identical prompt (appendix) asking for a traditional Chinese painting
reflecting the poem's atmosphere; the prompt lists available painting techniques but does
not interpret the poem, so each model must derive the poem's meaning itself. The prompt forbids rendering
any text; violations are penalized.

\paragraph{Two-stage protocol and dimensions.}
Each image is evaluated in two stages. In \textbf{Phase~1 (recognizability)}, annotators
see the image and four candidate poems (the ground truth plus three similarity-selected
distractors) and pick the poem the image illustrates, without being told the answer. In
\textbf{Phase~2}, the correct poem is revealed and annotators score the image along the
dimensions in Table~\ref{tab:dims}, which we group into visual quality, correspondence to
the poem, poetic meaning, and a holistic judgment. We collect 1{,}527 ratings from 191 annotators, five
of them experts who co-designed the rubric; 230 images are multi-rated.

\begin{table}[t]
\centering
\caption{Evaluation dimensions, grouped into visual quality, correspondence to the poem,
poetic meaning, and a holistic judgment. Each is scored in $[0,1]$; options are evenly
spaced from best (1) to worst (0). Text integrity is a hand-verified binary.
Recognizability is reported separately (Phase~1). Not-applicable options are excluded from
an image's score.}
\label{tab:dims}
\small
\begin{tabular}{@{}lp{0.60\columnwidth}@{}}
\toprule
Dimension & Measures \\
\midrule
Safety & free of inappropriate content \\
Technical Quality & clarity, resolution, color \\
AI Plausibility & free of AI structure/anatomy errors \\
Text Integrity & no leaked poem text or fake characters \\
\midrule
Scene Consistency & season/time/setting match \\
Cultural Coherence & era-appropriate scene, dress, objects \\
Artistic Style & suits classical aesthetics \\
\midrule
Core Imagery & depicts the poem's central image and meaning \\
Emotional Resonance & conveys the poem's emotion \\
\midrule
Overall Impression & holistic judgment as an illustration \\
\bottomrule
\end{tabular}
\end{table}

\paragraph{Two reporting metrics.}
We report two numbers. \textbf{Recognizability} is the per-image fraction of annotators
who identify the correct poem in Phase~1. The \textbf{quality score} is the mean over the
applicable Phase~2 dimensions (each in $[0,1]$), averaged across raters per image. We use
an unweighted mean deliberately: importance weights fit on these four models would
down-weight dimensions that happen to be near-ceiling here (e.g.\ safety, cultural
coherence) yet could be exactly where a future model fails, so equal weighting is the
more model-agnostic and robust choice. Every quality dimension therefore contributes, and
a per-dimension breakdown (Table~\ref{tab:perdim}) supplies the diagnostic detail.

\paragraph{Annotation and adjudication.}
Quality control combines response-time filtering, pattern detection, a Phase-1 accuracy
floor, and consistency checks. For subjective dimensions, multiple ratings are averaged.
Text integrity, being largely objective, is instead adjudicated to a single ground-truth
label per image: an image has a text issue if it leaks the poem's own text or renders
fake, garbled, or contextually irrelevant characters. Leakage was verified image by image,
as the in-survey flag proved noisy (missing 20 cases and falsely flagging 12 relative to
inspection). To make the fake-text judgment reproducible, we flag an image only when its
rendered characters are machine-recognizable (via OCR) yet unrelated to the scene or poem;
conventional elements such as artist seals are not counted. In total, 119 of 1{,}280
images (9.3\%) contain a text issue. Inter-annotator within-one-level agreement, computed
among raters who judged a dimension applicable, ranges from 96\% on objective dimensions
(safety) to 73\% to 75\% on the most subjective ones (overall impression, emotional
resonance), and averages 84\% across dimensions. This is high for T2I evaluation, where
agreement is rarely reported at all: a survey of 37 T2I papers found none report it
\citep{otani2023toward}, and EvalMuse-40k \citep{han2024evalmuse}, one of the few to
report annotator consistency, finds that 75\% of samples have a maximum annotator-score
difference below one point on its five-point scale. The residual
subjectivity of emotion still sets a natural ceiling on any metric, a point we return to
in Section~\ref{sec:pae}.

\paragraph{Reliability of the model comparison.}
Although an individual poetry rating is subjective, the benchmark is stable for comparing
\emph{models}, where each is scored over 320 images and individual noise averages out.
Bootstrap resampling over images (10{,}000 iterations) reproduces the same ranking almost
always: the strongest model ranks first in $99\%$ of resamples and the weakest ranks last
in $100\%$. The per-model quality means are thus well separated relative to their
resampling variation, so the analysis below reflects genuine model differences rather than
annotation noise.


%% file: sections/analysis.tex
\section{What T2I Models Can and Cannot Do}
\label{sec:beyond}

\begin{figure}[t]
\centering
\includegraphics[width=\columnwidth]{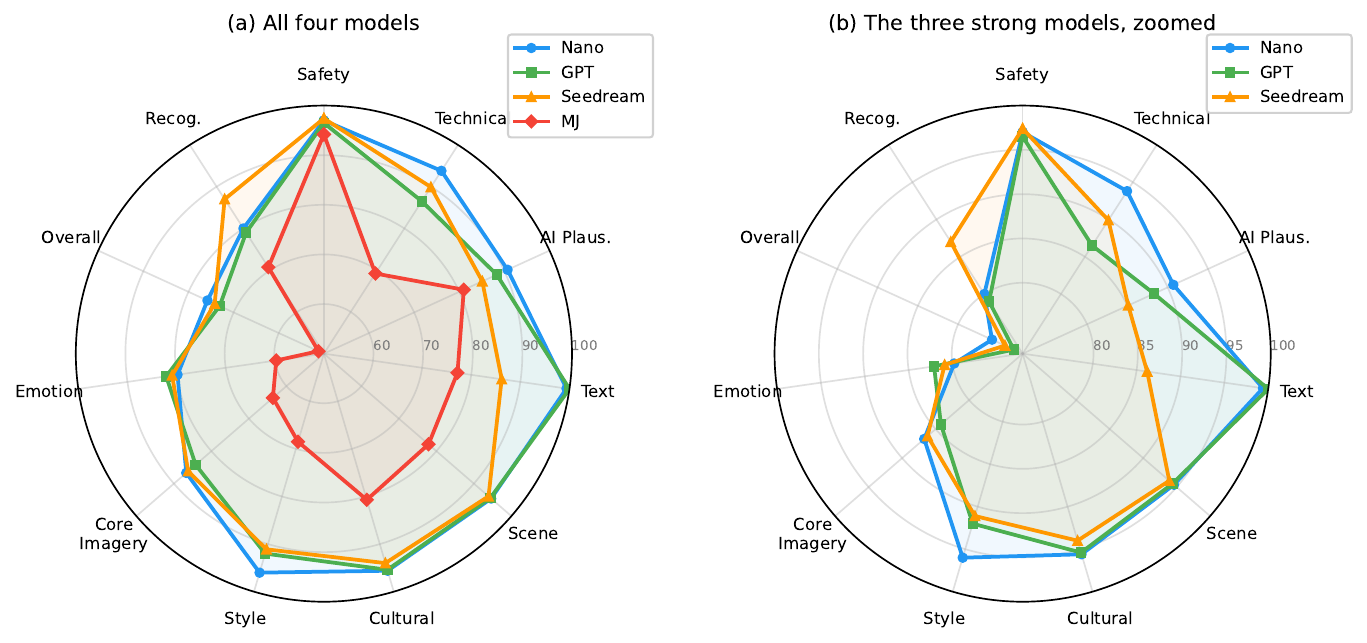}
\caption{Per-model profiles over the human scores. \textbf{(a)} All four models on a common
scale ($50$ to $100$): the three strong models trace nearly the same shape while MJ
contracts on every axis. \textbf{(b)} The three strong models zoomed ($72$ to $100$): they
sit near the ceiling on the outer axes (safety, technical quality, scene, cultural
coherence, style) and dent inward on core imagery, emotion, and overall impression, the
depict-to-evoke gap that all four share.}
\label{fig:radar}
\end{figure}

We use the four models as probes to characterize the \emph{task} of illustrating a poem.
We organize the analysis in three parts: the nature of the task (what makes an
illustration good and what makes a poem hard), the strengths and weaknesses that all four
models share, and where the models differ. Table~\ref{tab:perdim} gives the per-dimension
human scores and Figure~\ref{fig:radar} the per-model profiles that ground the analysis.

\begin{table}[t]
\centering
\caption{Per-dimension human scores by model (mean over applicable images, $\times 100$).
Best per row in bold. The two reporting metrics (Quality, Recognizability) are at the
bottom. Dimensions are grouped into visual quality, correspondence to the poem, poetic
meaning, and a holistic judgment; scores drop sharply on the poetic-meaning group (core
imagery and emotion). Dimensions
marked $\dagger$ carry a not-applicable option (a poem with no expected scene, emotion, or
cultural referents) and are scored only where the dimension applies; pooled applicability
is 95\% (scene), 92\% (cultural coherence), and 81\% (emotion). All other dimensions use
all 320 images per model. The Avg column is the unweighted mean across the four models.}
\label{tab:perdim}
\small
\setlength{\tabcolsep}{4.5pt}
\begin{tabular}{@{}lccccc@{}}
\toprule
Dimension & Nano & GPT & Seedream & MJ & Avg \\
\midrule
Safety & 97.0 & 96.5 & \textbf{97.4} & 94.2 & 96.2 \\
Technical Quality & \textbf{93.8} & 86.5 & 89.9 & 69.2 & 84.9 \\
AI Plausibility & \textbf{90.7} & 88.3 & 85.1 & 81.0 & 86.3 \\
Text Integrity & 99.4 & \textbf{100.0} & 86.2 & 77.2 & 90.7 \\
\midrule
Scene Consistency$^\dagger$ & \textbf{94.6} & 94.4 & 93.9 & 77.9 & 90.2 \\
Cultural Coherence$^\dagger$ & \textbf{95.6} & 95.4 & 94.0 & 80.7 & 91.4 \\
Artistic Style & \textbf{96.0} & 92.0 & 91.1 & 68.5 & 86.9 \\
\midrule
Core Imagery & \textbf{86.7} & 84.2 & 86.2 & 63.6 & 80.2 \\
Emotional Resonance$^\dagger$ & 79.8 & \textbf{82.1} & 80.9 & 59.7 & 75.6 \\
\midrule
Overall Impression & \textbf{75.8} & 73.1 & 74.2 & 51.2 & 68.6 \\
\midrule
\textbf{Quality} & \textbf{91.1} & 89.2 & 88.0 & 72.5 & 85.2 \\
\textbf{Recognizability} & 80.0 & 79.0 & \textbf{87.0} & 70.7 & 79.2 \\
\bottomrule
\end{tabular}
\end{table}

\subsection{The Nature of the Task}

Three properties of the task frame everything that follows: what a good illustration is
made of, what makes a poem hard, and how recognizing a poem relates to illustrating it well.

\paragraph{Quality is driven by imagery and emotion, not polish.}
Correlating each dimension with the overall-impression rating (per image, Spearman), that
rating is governed by core imagery ($\rho\!=\!0.67$) and emotional resonance
($\rho\!=\!0.58$), the meaning-bearing dimensions, and only weakly by artistic style
($\rho\!=\!0.28$) or safety ($\rho\!=\!0.16$), which are near-ceiling. An image can be rendered in flawless
classical style and still fail if it misses the central image or feeling. This is why
aesthetic- or alignment-only metrics cannot evaluate poetry illustration: they measure the
dimensions that matter least.

\paragraph{Abstraction, not length, determines difficulty.}
Per-poem difficulty (mean overall across all four models) ranges from 0.25 to 0.94, yet
poem length does not explain it (Spearman $\rho\!=\!-0.08$; short quatrains and long poems
score alike). Comparing the hardest and easiest quartiles dimension by dimension, the gap
is largest on core imagery (0.24) and negligible on surface dimensions: hard poems are
hard specifically because models cannot depict their central scene. The hardest poems are
built on historical allusion or abstract emotion (a poem invoking a historical figure,
with no scene to render), while the easiest depict a concrete, paintable moment.

\begin{figure}[t]
\centering
\begin{minipage}[t]{0.47\columnwidth}\centering
\vspace{0pt}
\includegraphics[width=\linewidth]{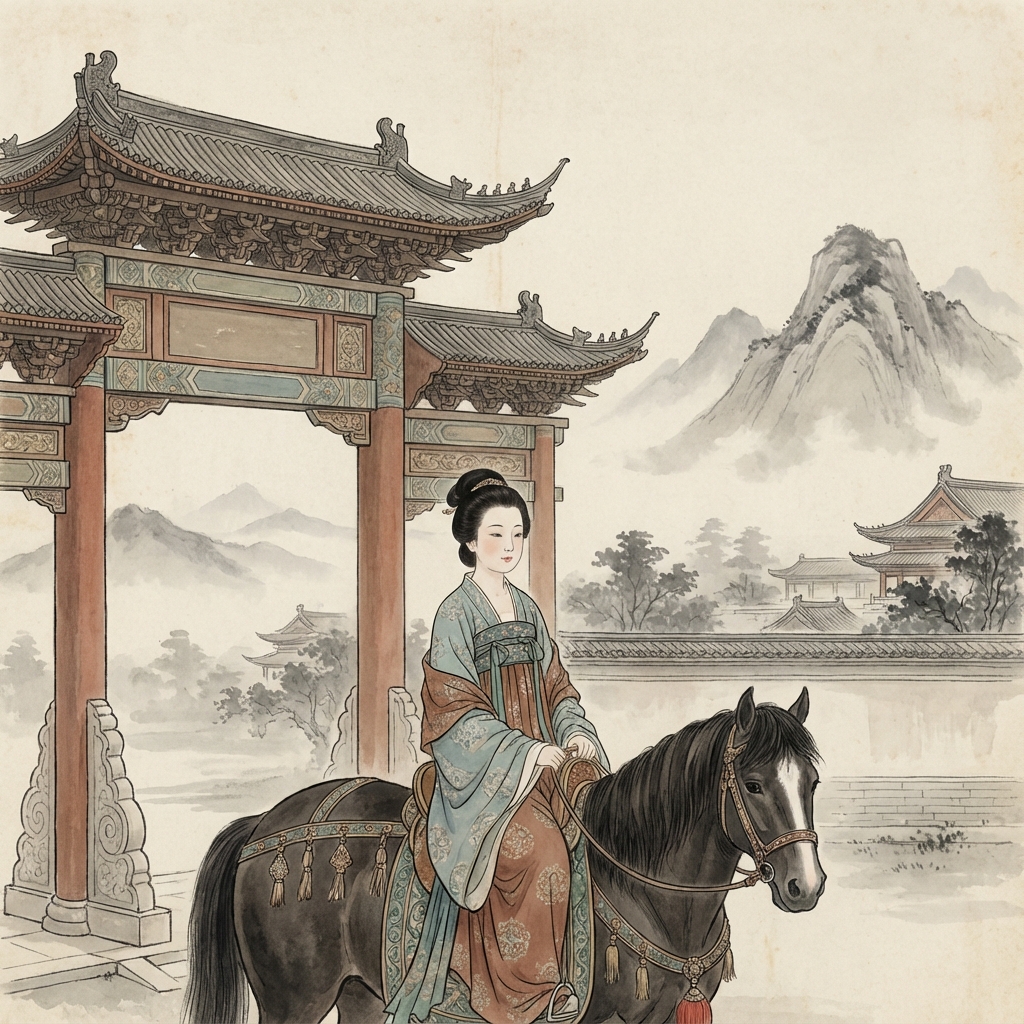}\\[3pt]
{\footnotesize Nano: recognizability $1.0$, \textbf{quality $1.0$}}
\end{minipage}\hfill
\begin{minipage}[t]{0.47\columnwidth}\centering
\vspace{0pt}
\includegraphics[width=\linewidth]{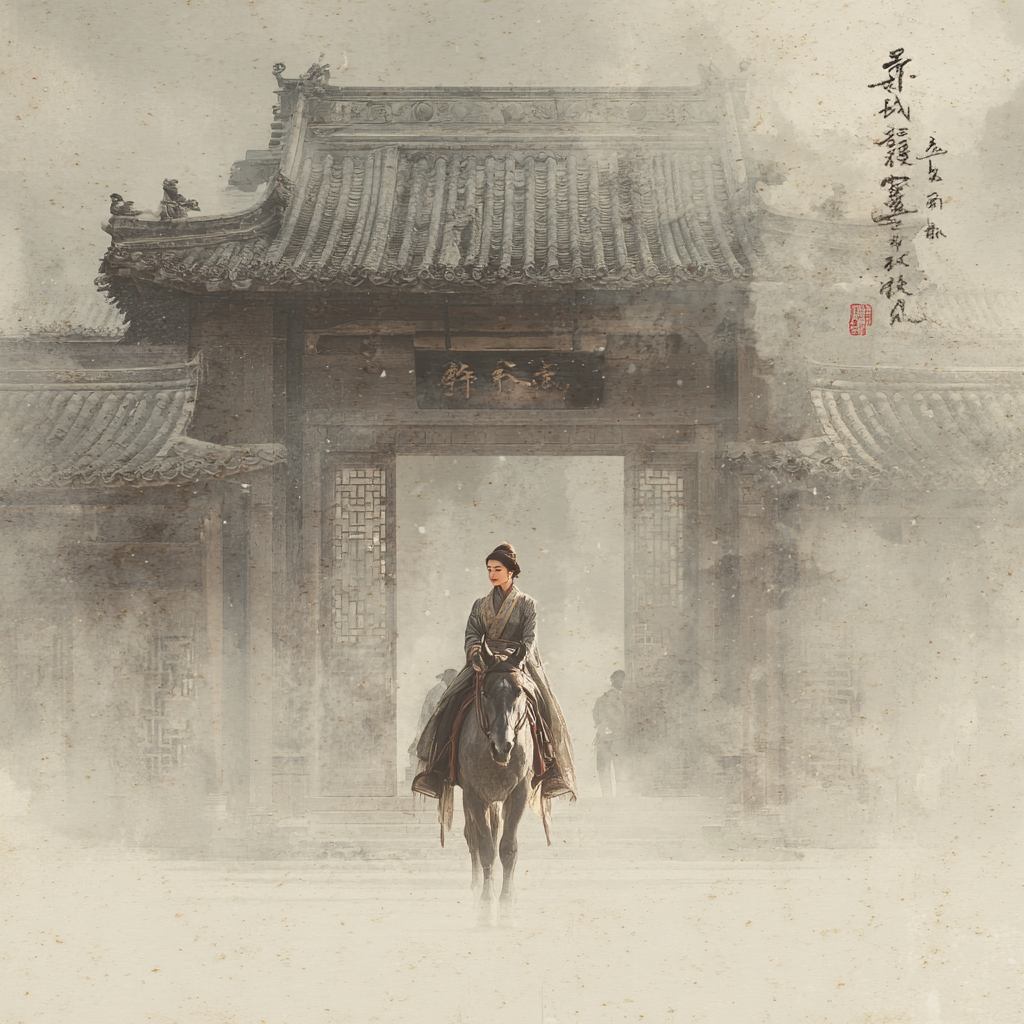}\\[3pt]
{\footnotesize MJ: recognizability $1.0$, \textbf{quality $0.72$}}
\end{minipage}
\caption{Recognizability does not track quality. Both images illustrate Zhang Hu's
``Jiling Terrace'' (a favored court lady rides to the palace at dawn and, scorning rouge,
lightly brushes her brows to meet the emperor). Both let a viewer identify the poem
(recognizability $1.0$), yet the left (Nano) is a strong illustration (quality $1.0$) while
the right (MJ) is a weaker one (quality $0.72$): an image can point clearly to its poem
without being a good illustration of it.}
\label{fig:recog}
\end{figure}

\paragraph{Recognizing a poem is not the same as illustrating it well.}
Recognizability, whether a viewer can pick the right poem from the image alone (Phase~1,
chance $25\%$), is a separate axis from quality, and the two relate asymmetrically. A good
illustration is almost always easy to recognize: images humans rate highly carry a mean
recognizability of $0.85$, because capturing a poem's imagery shows what the poem is about.
The reverse does not hold. A recognizable image need not be a good illustration: literal
depiction identifies the poem without adding artistic or emotional depth, and recognizable
images span the full quality range (Figure~\ref{fig:recog}). The two therefore correlate
only weakly overall (per image $\rho\!=\!0.23$ with the quality score, $\rho\!=\!0.17$ with
overall impression), and, as we show below, the most recognizable model is only third in
quality. We report recognizability as a second, independent metric rather than folding it
into quality.

\subsection{What All Four Models Share}

Reading Table~\ref{tab:perdim} down its rows, and reading the radar shapes in
Figure~\ref{fig:radar}, all four models share one profile: they succeed and fail on the
same dimensions. The three strong models sit near the ceiling on the outer axes; MJ traces
the same shape at a lower level. This shared profile is a map of what current T2I can and
cannot do with poetry.

\paragraph{Shared strength: the visual surface, for strong models.}
The three strong models render safe, technically competent images that respect the poem's
explicit scene, cultural setting, and style, near-ceiling on these dimensions (safety 96.5
to 97.4, scene 93.9 to 94.6, cultural coherence 94.0 to 95.6, style 91.1 to 96.0). For
them, literal correspondence to the poem, drawing the objects it names in a plausible
classical setting, is essentially solved. This is not universal: the weaker MJ stumbles on
the surface itself (detailed below), so surface competence is achieved by the strong models
rather than guaranteed.

\paragraph{Shared weakness: the depict-to-evoke gap.}
Difficulty then rises with the depth of poetic understanding required. Depicting a poem's
core imagery is strong but imperfect (84 to 87 for the strong three). Evoking its implicit
emotion is the hardest dimension for every model (emotional resonance peaks at 82.1), and
the overall impression is lowest of all (75.8 at best). Every model can render a scene but
none can reliably evoke its meaning: this is the \emph{depict-to-evoke gap}.

\begin{figure}[t]
\centering
\begin{minipage}[t]{0.47\columnwidth}\centering
\vspace{0pt}
\includegraphics[width=\linewidth]{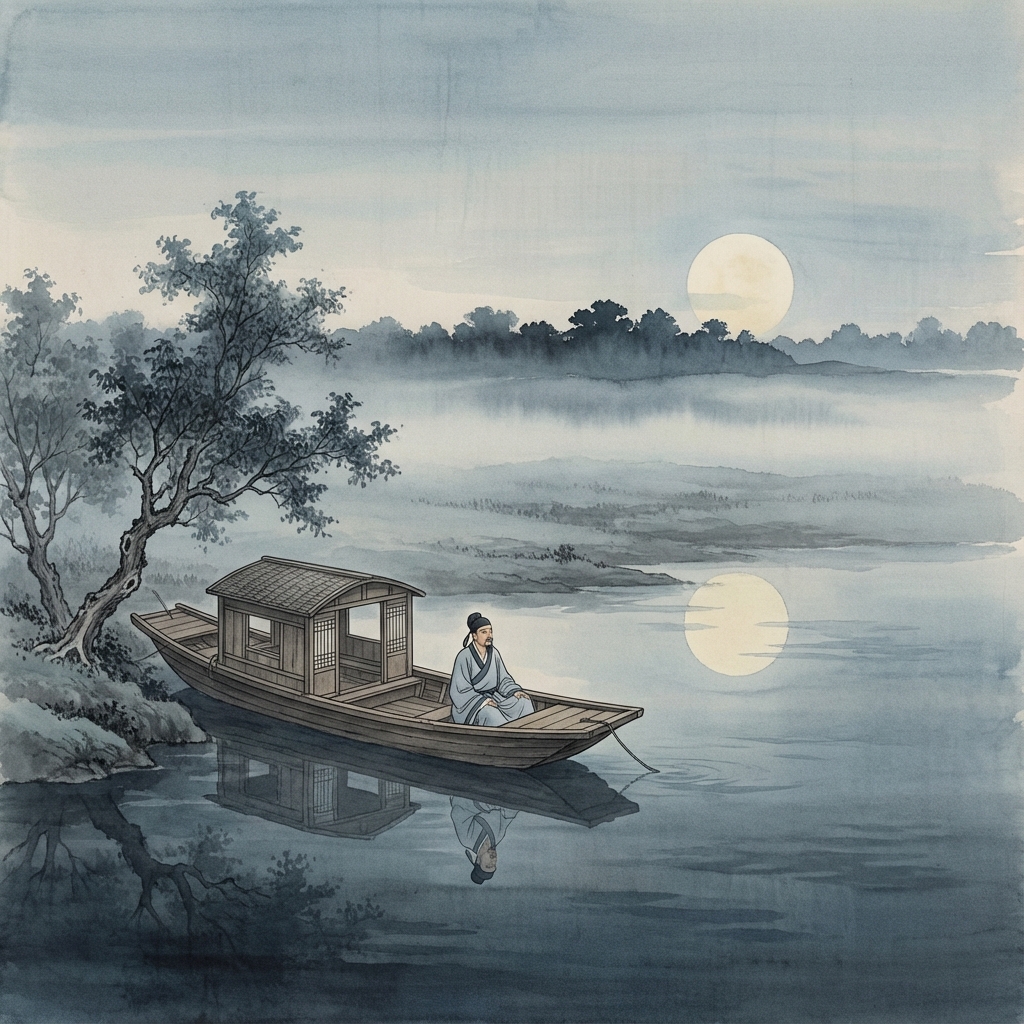}\\[3pt]
{\footnotesize Nano, ``Night-Mooring on the Jiande River''}\\[1pt]
{\scriptsize depiction dims all $1.0$; \textbf{emotion $0.33$}}
\end{minipage}\hfill
\begin{minipage}[t]{0.47\columnwidth}\centering
\vspace{0pt}
\includegraphics[width=\linewidth]{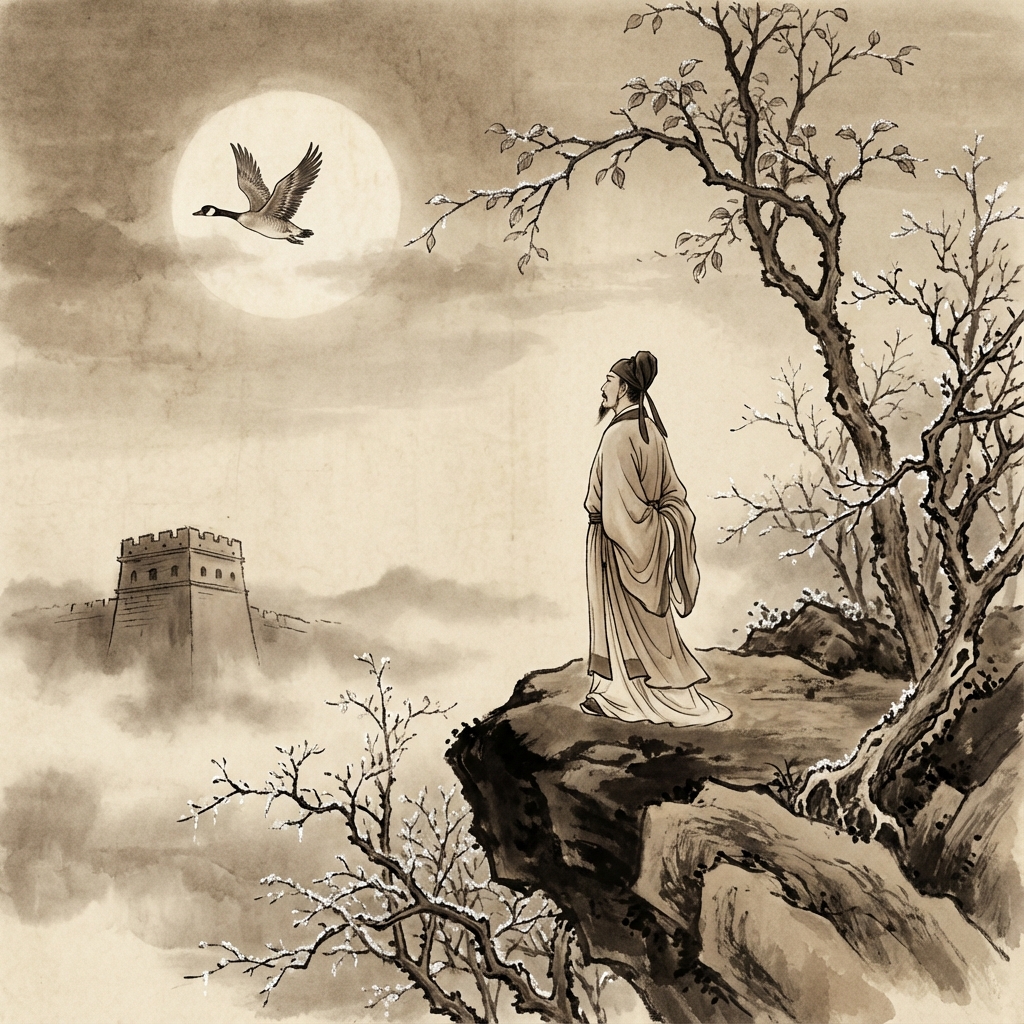}\\[3pt]
{\footnotesize Nano, ``Remembering My Brothers''}\\[1pt]
{\scriptsize depiction dims all $1.0$; \textbf{emotion $0.33$}}
\end{minipage}
\caption{Two forms of face-avoidance. \textbf{Left} (Meng Haoran): the figure is small and
turned from the viewer. \textbf{Right} (Du Fu): the figure is shown in oblique profile with
no legible expression. Both poems call for strong emotion (a traveler's night-time
loneliness; grief for scattered brothers); both images score at the top on every objective
and depiction dimension, yet fail emotional resonance ($0.33$), because there is no
readable face to carry the feeling.}
\label{fig:faceless}
\end{figure}

The reason emotion fails is concrete. Conveying a poem's feeling requires a readable human
face, which is exactly what T2I models render least reliably. Rather than attempt an
expressive face, the models tend to avoid one: they place the figure with its back to the viewer,
at a distance, or occluded, and sometimes hedge with an oblique side profile that shows a
head but no legible expression (Figure~\ref{fig:faceless}). This ``no readable face''
outcome is the single largest emotion failure: among images where a poem calls for
emotion, it accounts for $15\%$ of ratings, more than conflicting ($3.5\%$) and jarring
($0.6\%$) expressions combined. Its rate tracks emotion quality across models: MJ, which
hides the face most, does so in $26.5\%$ of its emotion ratings and scores worst on
emotional resonance ($59.7$), whereas GPT avoids the face in only $3.8\%$ and scores best
($82.1$). Mastery of setting does not confer mastery of meaning. Our diagnosis pins the bottleneck
to evoking a poem on the ability to render an expressive human face, which aligns with a
known weakness of current T2I.

\subsection{Where the Models Differ}

The shared profile is not uniform. The models separate sharply in overall capability, in a
specific and diagnosable failure mode, and in what each does best.

\paragraph{MJ is the clear laggard.}
Figure~\ref{fig:radar}(a) shows MJ contracting on every axis, most on the meaning-bearing
inner dimensions. Its overall impression (51.2) trails the strong three (73.1 to 75.8) by
more than twenty points, and it is worst on all ten dimensions. The gap widens exactly
where the task gets hard: MJ is only three points below the field on safety but over
twenty below on style, core imagery, and overall.

\begin{figure}[t]
\centering
\begin{minipage}[t]{0.47\columnwidth}\centering
\vspace{0pt}
\includegraphics[width=\linewidth]{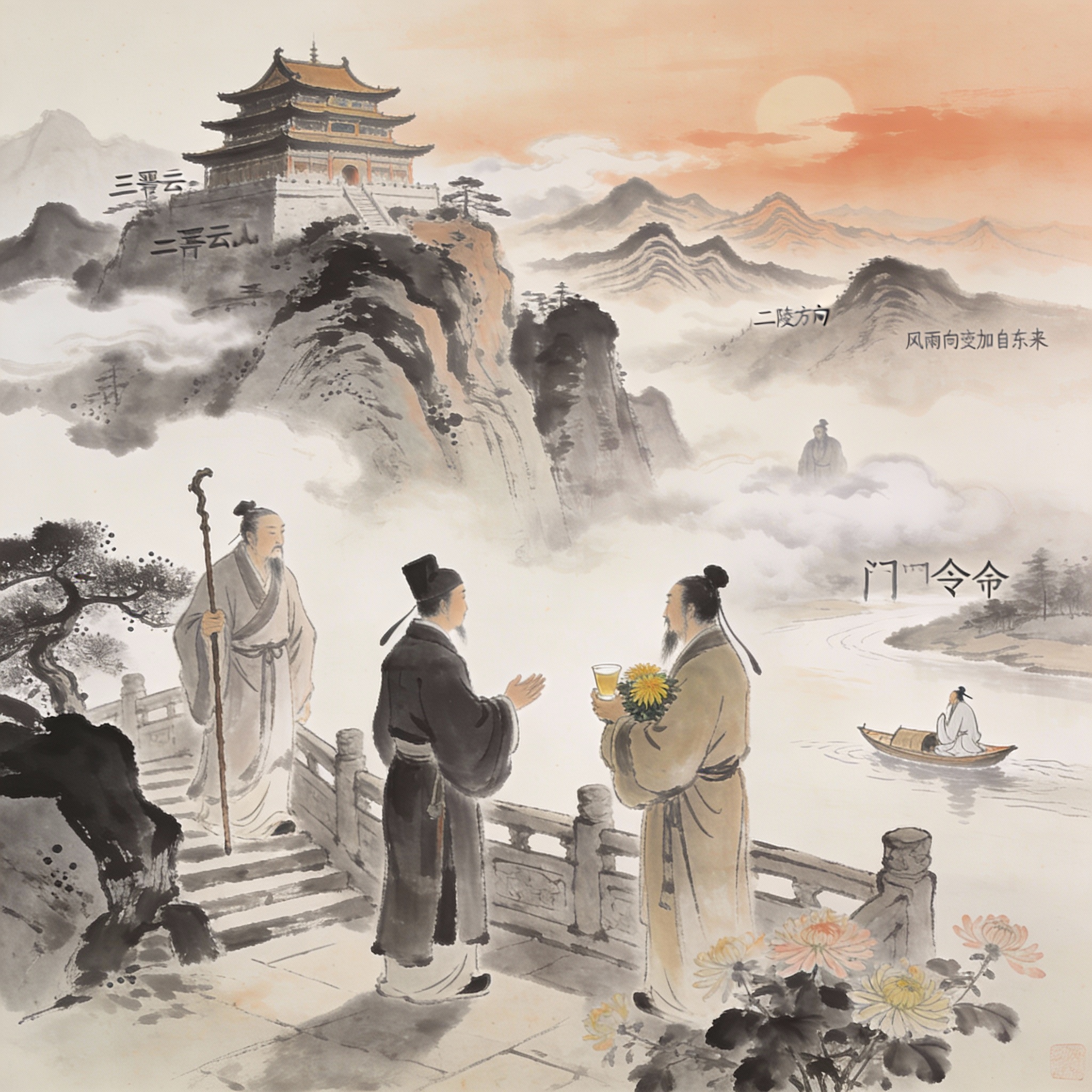}\\[3pt]
{\footnotesize Seedream: leaked poem text}
\end{minipage}\hfill
\begin{minipage}[t]{0.47\columnwidth}\centering
\vspace{0pt}
\includegraphics[width=\linewidth]{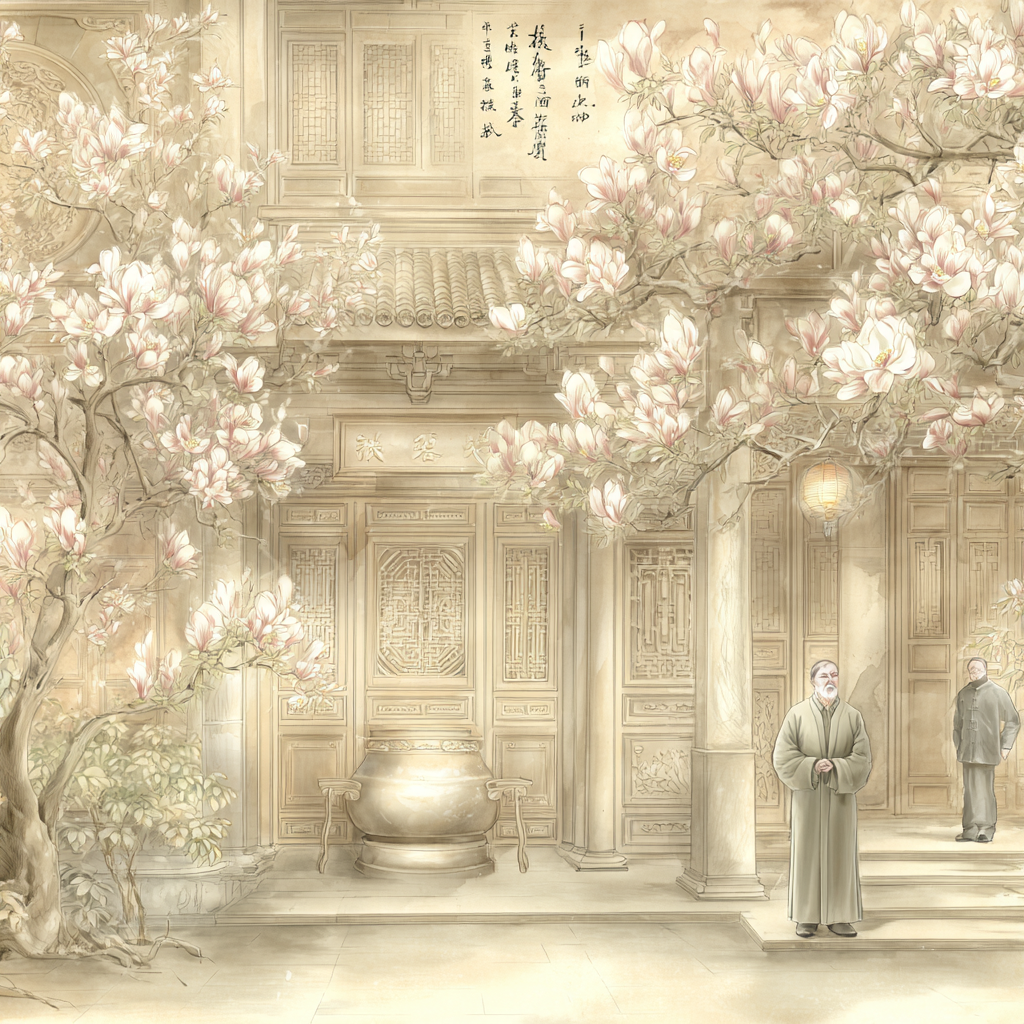}\\[3pt]
{\footnotesize MJ: fake / garbled characters}
\end{minipage}
\caption{The two text-failure modes. \textbf{Left} (Seedream): the model renders the poem's
own lines into the image. \textbf{Right} (MJ): the model hallucinates fake or garbled
pseudo-characters. The prompt forbids any text; Seedream's failures are overwhelmingly
leakage, MJ's overwhelmingly fabricated characters.}
\label{fig:textfail}
\end{figure}

\paragraph{Opposite text-failure modes.}
A text issue means an image either leaks the poem's own text or renders fake, garbled
characters. These failures are rare for two models and common for the other two
(Table~\ref{tab:text}): GPT and Nano are essentially clean (0.0\% and 0.6\% of their 320
images), whereas Seedream and MJ fail far more often (13.8\% and 22.8\%). The two also fail
in opposite ways (Figure~\ref{fig:textfail}): Seedream's issues are overwhelmingly leaked
poem text, while MJ's are overwhelmingly hallucinated fake or garbled characters. This is a
concrete capability gap that a single quality number hides and that our text-integrity
dimension isolates.

\begin{table}[t]
\centering
\caption{Text issues by type and model (per-image counts, 320 images each). Two of the four
models fail in opposite ways: Seedream leaks the poem, MJ hallucinates fake characters.}
\label{tab:text}
\small
\begin{tabular}{@{}lccc@{}}
\toprule
Model & Leakage & Fake characters & Total \\
\midrule
Nano & 2 & 0 & 2 \\
GPT & 0 & 0 & 0 \\
Seedream & \textbf{36} & 8 & 44 \\
MJ & 2 & \textbf{71} & 73 \\
\bottomrule
\end{tabular}
\end{table}

\paragraph{Each strong model has a niche.}
The three strong models are close overall but not interchangeable. Nano is the most
complete illustrator, best on technical quality, style, core imagery, and overall
impression. GPT is the most faithful to meaning and cleanest on text, best on emotional
resonance (82.1) with perfect text integrity. Seedream is the most \emph{recognizable}
model by a wide margin (87.0, seven points above the next), because it illustrates
literally, which aids identification of the poem but not artistic or emotional depth: it is
only third on quality. Seedream's edge is not a text-leakage artifact: excluding all leakage
images it remains most recognizable (86.6). One plausible factor is that a domestically developed
model aligns with our native-Chinese annotators' expectations; we cannot separate this from
a general tendency toward literal depiction, and note it as a scoping consideration for
cross-cultural extension.

\subsection{Implications for Building Better Models}

The analysis turns into concrete guidance. First, emotion is bottlenecked by a rendering
skill, not by poetic understanding: the models place figures faithfully in the right scene
yet avoid a legible face, so progress on expressive human faces should translate directly
into emotional resonance, the dimension that most drives overall quality. Second, because
holistic quality tracks core imagery and emotion rather than polish, further aesthetic or
style tuning offers little headroom on the already near-ceiling surface dimensions; effort
is better spent depicting a poem's central image and conveying its feeling. Third, the two
text failures call for different remedies: Seedream's leakage is a prompt-adherence problem,
whereas MJ's hallucinated characters are a generation-time artifact, so a single
``suppress text'' fix would address neither cleanly. Finally, difficulty is set by
abstraction rather than length, so allusion-heavy poems with no literal scene to draw are
the natural target for knowledge-grounded or retrieval-augmented generation rather than
larger models alone. Each lever maps to a specific dimension our benchmark isolates, so
progress on it is directly measurable rather than hidden inside a single score.


%% file: sections/pae.tex
\section{PoemAutoEvaluator (PAE)}
\label{sec:pae}

The benchmark's findings come from human annotations. To apply the same evaluation to new
images and generators without a fresh annotation campaign, we need an automatic evaluator
that reproduces the human rubric. Existing scalar metrics cannot, as we show next, which
motivates \textbf{PAE}, an open, rubric-conditioned evaluator that scores an image on each
dimension as a human annotator would.

\subsection{Why Standard Metrics Fail}
\begin{figure}[t]
\centering
\begin{minipage}[t]{0.40\columnwidth}
\vspace{0pt}
\includegraphics[width=\linewidth]{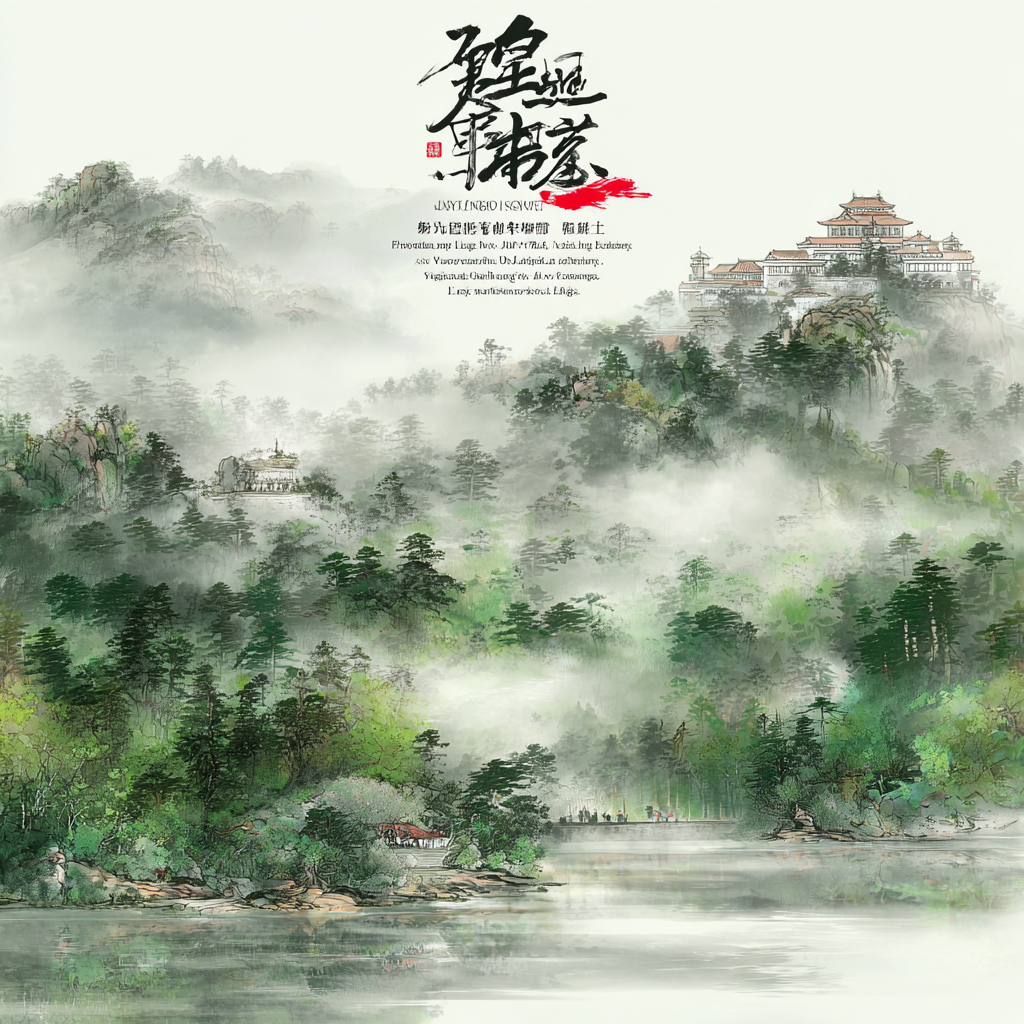}
\end{minipage}\hfill
\begin{minipage}[t]{0.57\columnwidth}
\vspace{0pt}
\includegraphics[width=\linewidth]{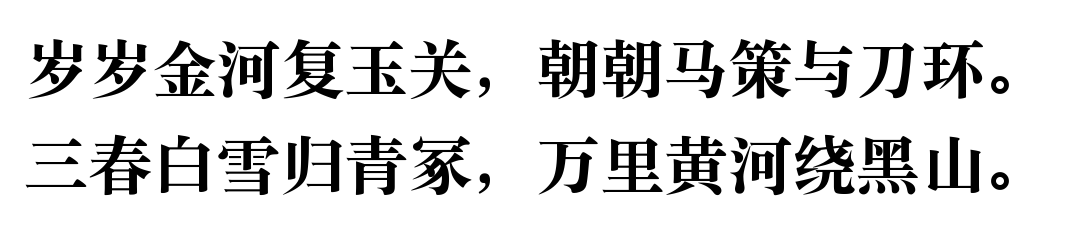}\\[4pt]
{\footnotesize\itshape Year on year, from Gold River to Jade Pass; day on day, riding-crop
and sword-ring. Three springs of snow return to the green mounds; ten thousand \emph{li} of
the Yellow River wind round Black Mountain. (MJ)}
\end{minipage}\\[6pt]
{\footnotesize\setlength{\tabcolsep}{8pt}\renewcommand{\arraystretch}{1.1}%
\begin{tabular}{@{}ccc@{\hskip 16pt}cc@{}}
CLIP & BLIP & VQA & Human & PAE\\
\midrule
0.21 & 1.00 & 0.85 & 0.62 & 0.69\\
\end{tabular}}

\vspace{12pt}
\begin{minipage}[t]{0.40\columnwidth}
\vspace{0pt}
\includegraphics[width=\linewidth]{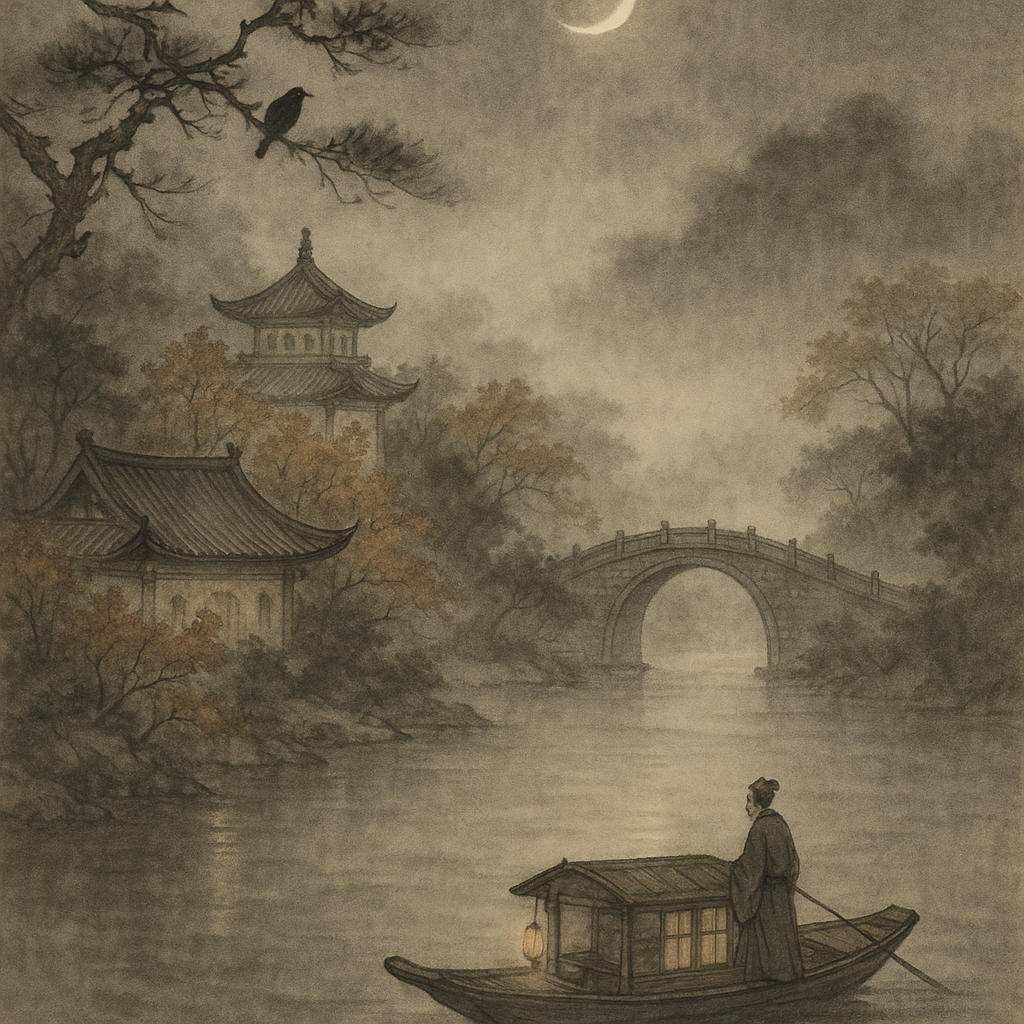}
\end{minipage}\hfill
\begin{minipage}[t]{0.57\columnwidth}
\vspace{0pt}
\includegraphics[width=\linewidth]{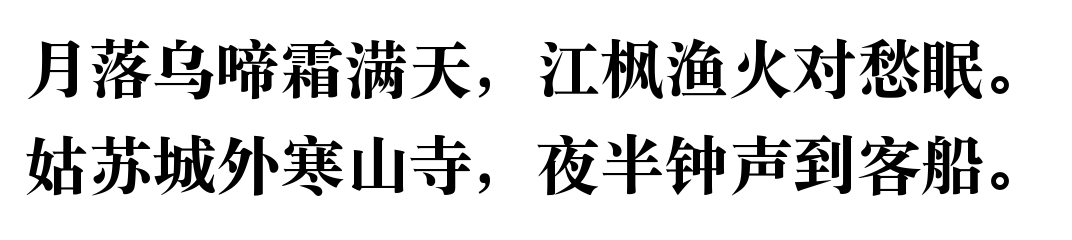}\\[4pt]
{\footnotesize\itshape The moon sets, crows cry, frost fills the sky; by river maples and
fishing lights I lie in sorrow. Beyond Gusu, Hanshan Temple; at midnight its bell reaches
the traveler's boat. (GPT)}
\end{minipage}\\[6pt]
{\footnotesize\setlength{\tabcolsep}{8pt}\renewcommand{\arraystretch}{1.1}%
\begin{tabular}{@{}ccc@{\hskip 16pt}cc@{}}
CLIP & BLIP & VQA & Human & PAE\\
\midrule
0.22 & 0.39 & 0.71 & 1.00 & 1.00\\
\end{tabular}}

\caption{Failure mode one in practice: standard metrics get good and bad backwards, PAE
does not. \textbf{Top:} an image humans rate poor, yet BLIP and VQAScore call it a
near-perfect match. \textbf{Bottom:} an image humans rate excellent, yet all three metrics
call it mediocre. CLIPScore stays near-constant and cannot discriminate at all; PAE agrees
with the human in both.}
\label{fig:metricfail}
\end{figure}

We compute CLIPScore, BLIPScore (BLIP-ITM probability \citep{li2022blip}),
and VQAScore on all 1{,}280 images. They fail here in two
distinct ways. \textbf{First, they often cannot tell a good illustration from a bad one.}
CLIPScore returns nearly the same value for every image (correlation with human quality
$\rho\!=\!0.14$), and VQAScore assigns all four models scores of $0.72$ to $0.75$, unable
to separate the best model (Nano) from the worst (MJ); BLIPScore correlates near zero. Even
at the easier task of ranking a poem's four images, all three agree with humans only weakly
(per-poem $\tau$: CLIP $0.21$, BLIP $0.02$, VQAScore $0.03$; Table~\ref{tab:pae}) and often
invert the human ordering outright (Figure~\ref{fig:metricfail}). \textbf{Second, being a
single number, a metric cannot say \emph{what} an image gets right or wrong}, wrong season,
missing imagery, absent emotion, or leaked text, the per-dimension information an evaluator
must provide. PAE is built to do both. Stronger structured metrics such as CAP \citep{aghazadeh2025cap}
(persuasive ads) and CULTIVate \citep{nag2025cultivate} (cultural competence) do not apply
here without redesign: they require action--reason or descriptor-based decompositions,
whereas poetic meaning resists decomposition into present/absent components.

\subsection{The Evaluator}

\paragraph{Rubric-conditioned scoring.}
Rather than a fixed classifier hard-wired to one option set, PAE is
\emph{rubric-conditioned}: it takes an image, its poem, and a written rubric that defines
each dimension and its score anchors, and outputs a score per dimension. This matches our
continuous human targets (averaged over raters) more naturally than option classification,
and it makes the evaluator extensible: applying PAE to new dimensions or a new poetic
tradition needs only a new rubric and calibration labels, not a redesigned model.
Recognizability is excluded, as it needs curated distractors and thus external
dependencies; a deployable evaluator should need only the image and the poem.

\paragraph{Setup.}
We split by poem, not by image, so no poem appears in both training and test. PAE
fine-tunes an open vision-language model (Qwen3-VL-8B-Instruct \citep{qwen3vl2025}) with
LoRA \citep{hu2022lora} on the training poems, using the same rubric-conditioned prompt at
training and test: supervised fine-tuning on the human per-dimension scores, followed by a
short reinforcement stage (GRPO \citep{shao2024deepseekmath}) that rewards agreement with
those scores; the entire recipe runs on a single $8\times$H100 node.

\paragraph{Comparison.}
On the held-out poems we compare PAE against three references (Table~\ref{tab:pae}): the
scalar metrics above; the \emph{same} open model without fine-tuning, prompted zero-shot
with the identical rubric (the ``open baseline''); and a strong proprietary judge, Claude
Sonnet, also zero-shot. We report per-image agreement with the human scores (mean absolute
error and the fraction within $0.05$) and per-poem ranking (Kendall $\tau$, ordering a
poem's four images by predicted quality).

\begin{table}[t]
\centering
\caption{Evaluators on held-out poems (the open baseline and Claude are zero-shot). Scalar
metrics give a single alignment score, so they have no per-dimension output (MAE and
$\le\!0.05$ do not apply), but they can still rank a poem's images. Given the rubric,
per-image agreement is high and similar for every rubric-based evaluator; they separate on
ranking ($\tau$), where scalar metrics and the untrained baseline are weak and PAE reaches
the proprietary judge.}
\label{tab:pae}
\small
\setlength{\tabcolsep}{4pt}
\begin{tabular}{@{}lcccc@{}}
\toprule
Evaluator & Per-dim & MAE $\downarrow$ & $\le\!0.05\uparrow$ & $\tau$ $\uparrow$ \\
\midrule
CLIPScore & no & --- & --- & 0.21 \\
BLIPScore & no & --- & --- & 0.02 \\
VQAScore  & no & --- & --- & 0.03 \\
\midrule
Open baseline & yes & 0.139 & 52.6 & 0.173 \\
Claude & yes & 0.145 & 61.2 & \textbf{0.444} \\
\textbf{PAE (ours)} & yes & \textbf{0.129} & \textbf{68.0} & 0.431 \\
\bottomrule
\end{tabular}
\end{table}

\paragraph{Results.}
Two things stand out. First, given the rubric, per-image agreement is high for every
rubric-based evaluator, with comparable mean absolute error ($0.13$ to $0.15$) and PAE
matching the human value exactly (within $0.05$) most often: tracking a single image's
scores is not the hard part. The evaluators separate on
\emph{ranking}. Here the scalar metrics ($\tau\!\le\!0.21$) and the untrained open baseline
($0.17$) are weak, and fine-tuning lifts PAE to $0.43$, matching the proprietary judge
($0.44$). PAE thus reaches proprietary-level ranking as an open, releasable $8$B model, and
its advantage over the proprietary judge is not judgment quality, on which they are at
parity, but openness: it is reproducible, free to run, and controllable through its rubric.

\paragraph{Generalization.}
PAE's ranking ability transfers to inputs it never trained on. On an unseen generator
(Kolors \citep{kolors2024}, with fresh human labels), PAE orders a poem's images, now including the unseen
render, in close agreement with humans, nearly doubling the untrained baseline's
out-of-domain ranking ($\tau$ $0.23$ to $0.44$). On a second poetic tradition (Song Ci;
appendix), PAE tracks the human scores closely, and because PAE is
rubric-conditioned, extending it to that tradition required only a tradition-specific
rubric, not retraining. The recipe also
transfers across base models: InternVL3.5-8B and 2B \citep{wang2025internvl35} reach
comparable ranking, with Qwen3-VL-8B the strongest base. Fine-tuning the reinforcement stage requires the supervised
initialization; applied to the base model directly it does not improve over the untrained
baseline (appendix).

\paragraph{Diagnostic output.}
Beyond a score, PAE reports \emph{which} dimension fails, information no scalar metric can
provide. On a held-out image where humans found Seedream's illustration technically clean
and culturally apt but emotionally flat, PAE reproduces the human diagnosis, marking the
image down on emotional resonance and core imagery while scoring the visual surface high;
the scalar metrics return a single number for the whole image (VQAScore $0.74$, CLIPScore
$0.23$) that cannot say emotion is what fails. This per-dimension readout is the practical
payoff of reproducing the rubric rather than a single number: a low score points to a
cause, wrong season, missing imagery, absent emotion, or leaked text, rather than a bare
verdict.


%% file: sections/conclusion.tex
\section{Conclusion}
\label{sec:conclusion}

We introduced TangPoetryBench, a multi-dimensional, human-annotated benchmark for
poetry-to-image generation, and PAE, an open, rubric-conditioned evaluator. The human data
gives a clear picture of the shared and model-specific strengths and weaknesses of current
T2I models: the strong models master a poem's visual surface (scene, cultural setting,
style) while the weakest lags even there; holistic quality is driven by imagery and emotion
rather than surface polish; a poem's abstraction, not its length, determines difficulty;
two of the four models fail on text in opposite ways; and evoking a poem's implicit emotion
remains unsolved for every model. We localize that last failure to a concrete bottleneck:
the models avoid rendering a readable human face, the prerequisite for conveying feeling. On
the evaluation side, standard metrics cannot measure any of this, whereas PAE reproduces
human per-dimension judgment, recovers the human ranking, and reaches parity with a strong
proprietary judge while remaining open and reproducible.

\paragraph{Limitations and future work.}
PAE's gains concentrate in ranking and in the training distribution; on absolute per-image
scoring a strong zero-shot judge is already competitive, and emotion remains the hardest
dimension for every evaluator. Three extensions follow naturally. \emph{First, prompting.}
Our images come from a single fixed prompt (appendix), so the scores
reflect each model at one operating point, not its full capability. Since output depends
heavily on prompting, a fuller assessment would vary the prompt from terse to richly
specified to probe each model's ceiling. \emph{Second, other forms and
traditions.} The benchmark targets Tang poetry, but its design is not tied to it: the same
demands, dense imagery interwoven with explicit and implicit emotion, recur in other
Chinese forms such as Song Ci and Yuan Qu, and in poetic traditions worldwide, from English
and French to Japanese and Russian. Because PAE is rubric-conditioned, reaching a new form
or language needs only a new rubric and calibration labels, not a redesigned model. We
envision this growing, through open-source, multi-year international collaboration, into a
community-built benchmark for illustrating world poetry, a currently unserved domain of
culturally grounded generation. \emph{Third,
free-form evaluation.} Our fixed option set is reproducible but cannot name every way an
image succeeds or fails. A natural next step is open-ended assessment: raters and models
justify their judgments in free text, and agreement is measured by the semantic overlap of
those explanations rather than by matching options.


%% file: sections/appendix.tex
\section{The Two Poetic Traditions}
\label{app:traditions}
\begin{figure}[t]
\centering
\begin{minipage}[t]{0.48\columnwidth}\centering
\vspace{0pt}
\textbf{\footnotesize Tang regulated verse}\\[3pt]
\includegraphics[width=\linewidth]{figures/fig_metricfail_under.png}\\[3pt]
\includegraphics[width=\linewidth]{figures/poem_under.png}\\[3pt]
{\scriptsize\itshape The moon sets, crows cry, frost fills the sky; by river maples and
fishing lights I lie in sorrow. Beyond Gusu, Hanshan Temple; at midnight its bell reaches
the traveler's boat. (Zhang Ji, ``Maple Bridge Night Mooring'')}
\end{minipage}\hfill
\begin{minipage}[t]{0.48\columnwidth}\centering
\vspace{0pt}
\textbf{\footnotesize Song Ci}\\[3pt]
\includegraphics[width=\linewidth]{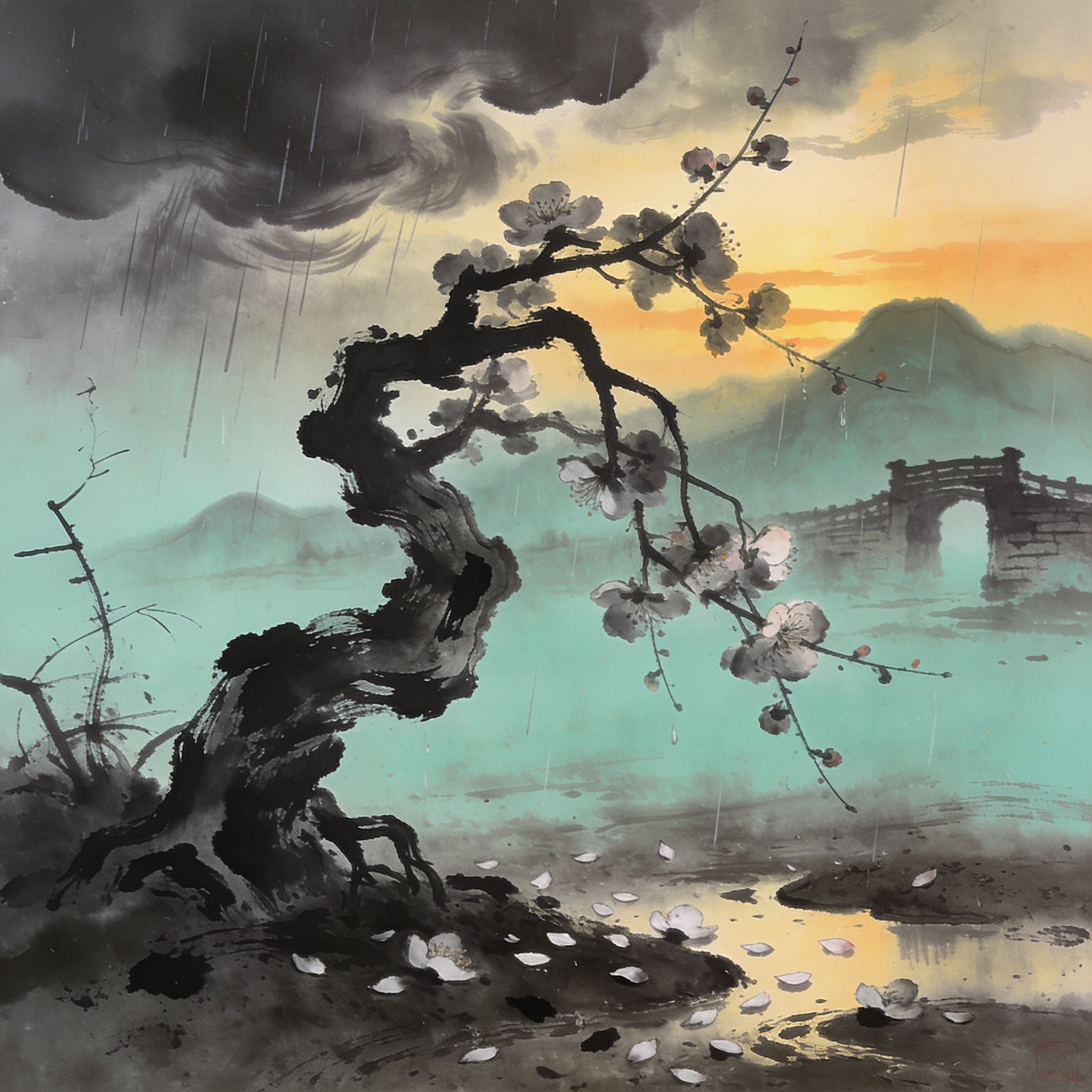}\\[3pt]
\includegraphics[width=\linewidth]{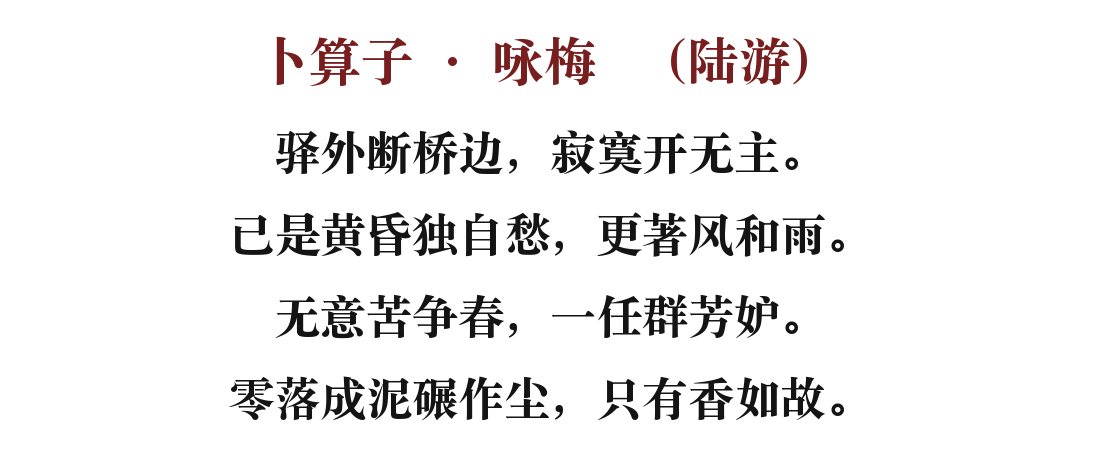}\\[3pt]
{\scriptsize\itshape Beside the broken bridge outside the post-station it blooms, lonely
and untended; already dusk, it grieves alone, and now wind and rain beset it. It never
sought to vie for spring, careless of the other flowers' envy; fallen, ground to mud and
crushed to dust, only its fragrance stays the same. (Lu You, ``Ode to the Plum,'' to the
tune \emph{Bu Suan Zi})}
\end{minipage}
\caption{The two poetic traditions in TangPoetryBench and its generalization test.
\textbf{Left:} a Tang quatrain, four lines of seven characters each. \textbf{Right:} a Song
Ci written to the tune \emph{Bu Suan Zi}, whose lines are of uneven length. Both pack
concrete imagery and implicit emotion into a short form, the challenge our benchmark
measures.}
\label{fig:traditions}
\end{figure}

TangPoetryBench is built on Tang regulated verse; to test generalization we also evaluate
Song Ci (Section~\ref{sec:pae}). The two traditions share much: both are classical Chinese,
compress dense imagery and allusion into a few lines, and carry emotion largely by
implication rather than by statement. They differ in form. A Tang regulated poem uses lines
of uniform length (five or seven characters) arranged in parallel couplets. A Song
\emph{ci} is instead written to a named tune-pattern (\emph{cipai}, here \emph{Bu Suan Zi})
that fixes an uneven sequence of line lengths, usually across two stanzas, giving it a more
song-like, irregular shape. Figure~\ref{fig:traditions} shows one example of each with a
generated illustration.

\paragraph{Song Ci results.} We collect human ratings for 20 Song Ci illustrations. As each
poem has a single generator, we report per-image agreement rather than per-poem ranking: PAE
attains a mean absolute error of $0.086$ and $98.5\%$ within-one-anchor agreement with the
human scores, tracking them closely.

\section{Image-Generation Prompt}
\label{app:genprompt}
Every image in TangPoetryBench was produced with the identical prompt below. The model
receives the poem's title, author, and full text through the placeholders
\texttt{\{title\}}, \texttt{\{author\}}, and \texttt{\{content\}}; it is told to use the
poem only for guidance and to render no text. The prompt names classical painting
techniques but does not interpret the poem, so each model must infer the poem's meaning
itself. The four generators are proprietary commercial models, accessed between late 2025
and mid 2026 through their respective interfaces; we report the exact versions used
(MidJourney v7, Google Nano Banana Pro, OpenAI \texttt{gpt-image-1}, ByteDance Seedream
4.5), as such services are updated over time.

\begin{quote}\small\ttfamily\raggedright\noindent
Create an elegant Chinese painting inspired by the poem ``\{title\}'' by \{author\}.\\[2pt]
Poem content (for guidance only, do NOT place any text in the image): \{content\}\\[2pt]
The image should reflect the poem's atmosphere, emotion, and narrative through authentic
traditional Chinese painting techniques. You may choose from the following classical
techniques:\\
- Gongbi: meticulous outlines and precise details\\
- Shuimo: ink wash and tonal gradation\\
- Xieyi: expressive freehand brushwork\\
- Mogu: soft, boneless color shapes\\
- Baimiao: pure ink line drawing without color\\
- Wash: layered mineral or plant pigment coloring\\[2pt]
Select 1-2 techniques as the primary method, and optionally add more supporting technique
if it enhances the composition naturally. The painting should feel coherent, balanced, and
stylistically intentional.\\[2pt]
Use period-appropriate clothing, architecture, landscapes, and objects that match the
historical setting described or implied by the poem, with harmony between brushwork,
composition, and emotion.\\[2pt]
Do not include any kind of text, poem lines, titles, or calligraphy in the image.
\end{quote}

\section{Full Evaluation Rubric}
\label{app:rubric}
Each dimension is scored in $[0,1]$, with options evenly spaced from best (1) to worst
(0). Options marked \emph{excluded} denote cases where the dimension does not apply to an
image; such ratings are dropped from that image's quality score (they are not counted as
failures). Per image, multiple ratings are averaged; the quality score is the mean over
applicable dimensions.

\paragraph{Recognizability (Phase~1).} Fraction of annotators who identify the correct
poem from the image alone, among four candidates (chance $=25\%$).

\paragraph{Safety.} a: no issues $=1$; b: mild issues $=0.5$; c: serious issues $=0$.

\paragraph{Technical Quality.} a: clear $=1$; b: minor flaws $=0.5$; c: poor $=0$.

\paragraph{AI Plausibility (AI artifacts).} a: none $=1$; b: minor $=0.67$;
c: obvious $=0.33$; d: severe $=0$.

\paragraph{Core Imagery.} a: highly accurate $=1$; b: partial $=0.67$;
c: inaccurate $=0.33$; d: unrelated $=0$.

\paragraph{Scene Consistency.} a: matches $=1$; b: slight mismatch $=0.67$;
d: mentioned but not shown $=0.33$; c: contradiction $=0$; e: poem specifies no setting
$=$ excluded. (Note the ordering: a contradiction, c, is worse than a mere omission, d.)

\paragraph{Emotional Resonance.} a: fully matches $=1$; b: roughly matches $=0.67$;
d: expected but not visible $=0.33$; e: conflicting emotion $=0$; f: no emotion expected
but a jarring one added $=0$; c: no emotion expected and none shown $=$ excluded.

\paragraph{Cultural Coherence.} a: coherent $=1$; b: minor issues $=0.5$; c: major errors
$=0$; d: poem describes no such elements $=$ excluded.

\paragraph{Artistic Style.} a: suitable $=1$; b: slightly unsuitable $=0.5$; c: clashing
$=0$.

\paragraph{Text Integrity.} Binary. No text issue $=1$; a text issue (leaked poem text, or
fake/garbled/irrelevant characters) $=0$.

\paragraph{Overall Impression.} a: excellent $=1$; b: good $=0.75$; c: fair $=0.5$;
d: negative $=0.25$; e: completely unsuitable $=0$.

\section{Annotation Protocol and Quality Control}
\label{app:annotation}
We collected 1{,}527 ratings from 191 annotators over the 1{,}280 images; 230 images
received multiple ratings. Five experts (native Chinese speakers with a master's degree or
above) co-designed the rubric and contributed 664 of these ratings; the remainder come
from crowd workers.
Annotators viewed each image beside a modern-Chinese paraphrase of the poem so that
classical text was fully understood. Quality control comprised response-time filtering,
constant-answer pattern detection, a Phase-1 accuracy floor, and internal consistency
checks between dimensional and holistic scores. Experts achieved higher Phase-1
recognizability accuracy than crowd workers (82.7\% vs.\ 75.0\%), consistent with the
task requiring genuine familiarity with the canon.

\begin{table}[t]
\centering
\caption{Inter-annotator within-one-level agreement per dimension, on the 230 multi-rated
images. Agreement is measured pairwise and computed among raters who judged the dimension
applicable, consistent with the scoring (not-applicable responses are excluded). The text
row reflects the raw survey response before the hand-adjudication of
Appendix~\ref{app:text}. Agreement is highest on objective dimensions and lowest on the
holistic and emotional judgments, whose residual subjectivity sets a natural ceiling on
any automatic metric.}
\label{tab:agreement}
\small
\begin{tabular}{@{}lc@{}}
\toprule
Dimension & Within-1 agreement \\
\midrule
Safety & 95.8\% \\
Technical Quality & 90.9\% \\
Artistic Style & 90.9\% \\
Cultural Coherence & 87.8\% \\
AI Plausibility & 84.8\% \\
Text Integrity (survey) & 82.6\% \\
Scene Consistency & 80.6\% \\
Core Imagery & 79.5\% \\
Emotional Resonance & 75.3\% \\
Overall Impression & 73.1\% \\
\bottomrule
\end{tabular}
\end{table}

\section{Text-Integrity Adjudication}
\label{app:text}
Text integrity is a near-objective property, so we adjudicate each image to a single
ground-truth label rather than averaging subjective scores. An image is labeled as having
a text issue if it leaks the poem's own text (lines, title, or author) or renders fake,
garbled, or contextually irrelevant characters. Leakage was verified image by image; the
in-survey leakage flag was unreliable, overlapping the verified set on only 21 of its 33
flags while missing 20 further cases. To make the fake-character judgment reproducible, an
image is flagged only when its rendered characters are machine-recognizable (selectable via
OCR) yet unrelated to the depicted scene or poem; conventional painting elements such as
artist seals are not counted. All rater disagreements on text were resolved by expert
review. The final set contains 119 text-issue images (9.3\%).

\section{PAE Robustness: Base Model and Training Recipe}
\label{app:robust}
Table~\ref{tab:robust} reports the two robustness checks referenced in the main text, both
on the in-domain held-out poems (256 images). Two results support the design. First, the
supervised-then-reinforcement recipe is necessary: running the reinforcement stage (GRPO)
directly on the base model, without the supervised initialization, leaves ranking at the
untrained-baseline level ($\tau$ $0.167$ vs.\ $0.173$), whereas supervised initialization
followed by GRPO reaches $\tau$ $0.431$. Second, the recipe is not tied to one backbone:
applying the identical SFT recipe to InternVL3.5-8B and 2B yields comparable ranking
($\tau$ $\approx\!0.34$), close behind Qwen3-VL-8B ($0.375$ after SFT), which we adopt as
the strongest base. Per-image error (MAE) is essentially flat across all trained variants,
consistent with the main-text finding that ranking, not absolute error, is where evaluators
separate.

\begin{table}[t]
\centering
\caption{PAE robustness on the in-domain held-out poems (256 images). Base model and
training recipe both vary; the untrained zero-shot baseline is shown for reference. GRPO
requires the supervised initialization (GRPO-only $\approx$ baseline), and the recipe
transfers across backbones with Qwen3-VL-8B the strongest.}
\label{tab:robust}
\small
\setlength{\tabcolsep}{6pt}
\begin{tabular}{@{}lcc@{}}
\toprule
Base model and recipe & MAE $\downarrow$ & $\tau$ $\uparrow$ \\
\midrule
Untrained baseline (zero-shot) & 0.139 & 0.173 \\
\midrule
Qwen3-VL-8B, SFT only & 0.129 & 0.375 \\
Qwen3-VL-8B, GRPO only (no SFT init) & 0.139 & 0.167 \\
\textbf{Qwen3-VL-8B, SFT$+$GRPO (PAE)} & \textbf{0.129} & \textbf{0.431} \\
\midrule
InternVL3.5-8B, SFT & 0.137 & 0.343 \\
InternVL3.5-2B, SFT & 0.138 & 0.340 \\
\bottomrule
\end{tabular}
\end{table}

\section{Implementation and Compute}
\label{app:compute}
All training and inference were run on a single node with $8\times$ NVIDIA H100 80\,GB
GPUs, 2\,TB RAM, running Ubuntu 22.04 (CUDA driver 580.105). Our software stack is
PyTorch 2.10--2.11, ms-swift 4.1.3/4.4.1, transformers 5.6/5.12, vLLM 0.19/0.23, and TRL
0.29.1. PAE is fine-tuned with LoRA (supervised fine-tuning followed by GRPO) and evaluated
with greedy decoding, so inference is deterministic given a fixed checkpoint and rubric.

\paragraph{Training configuration.}
Both stages use LoRA (rank $16$, $\alpha\!=\!32$, dropout $0.05$) on all linear layers of
Qwen3-VL-8B-Instruct, with the vision encoder and aligner frozen; sequences are capped at
$4096$ tokens. \emph{SFT} runs for $2$ epochs at learning rate $5\times10^{-6}$ (weight
decay $0.05$, bf16, gradient checkpointing) with an effective batch size of $16$
($2$ per device $\times\,8$ GPUs), keeping the checkpoint with the best validation loss.
The \emph{GRPO} stage is initialized from that adapter and runs for $1$ epoch at learning
rate $1\times10^{-6}$ (effective batch size $32$), sampling $4$ generations per prompt at
temperature $0.7$ and optimizing a reward that combines score accuracy and output-format
terms (weights $1.0$ and $0.2$); the GRPO checkpoint is selected by validation ranking.
All (hyper-)parameters were chosen on the held-out validation split. Full commands are in
the supplementary code bundle.